\pdfoutput=1 
\documentclass[lettersize,journal]{IEEEtran}
\usepackage{amsmath,amsfonts}
\usepackage{algorithm}
\usepackage{array}
\usepackage{textcomp}
\usepackage{stfloats}
\usepackage{url}
\usepackage{verbatim}
\usepackage{graphicx}
\usepackage{cite}
\usepackage{mathpazo,amsmath,amssymb,epsfig,enumerate,bbm,calc,color,ifthen,capt-of} 

\usepackage{color, graphicx}
\usepackage[scaled]{helvet} 
\usepackage{booktabs, multirow}
\usepackage{tabularx}
\usepackage[table]{xcolor}
\usepackage{subcaption}
\usepackage{balance}
\usepackage{bbding}
\usepackage{multirow}
\usepackage{threeparttable}
\usepackage{hyperref}
\usepackage[justification=raggedright]{caption}
\usepackage{algorithm}
\usepackage[noend]{algpseudocode}
\usepackage{multirow}

\title{\LARGE \bf
FSNet: Redesign Self-Supervised MonoDepth for Full-Scale Depth Prediction for Autonomous Driving 
}

\author{Yuxuan Liu, ~\IEEEmembership{Student Member, ~IEEE}, Zhenhua Xu, ~\IEEEmembership{Student Member, ~IEEE}, Huaiyang Huang, ~\IEEEmembership{Student Member, ~IEEE}, Lujia Wang, ~\IEEEmembership{Member, ~IEEE}, and Ming Liu, ~\IEEEmembership{Senior Member,~IEEE}
\thanks{*This work was supported by the National Natural Science Foundation
of China (Grant No. U1713211), the Research Grant Council of Hong
Kong SAR Government, China, under Project No. 11210017, and No.
21202816, and Shenzhen Science, Technology and Innovation Comission
(SZSTI) JCYJ20160428154842603, awarded to Prof. Ming Liu.} 
\thanks{The authors are with the Robotics and Multi-Perception Laborotary, 
Department of Electronic and Computer Engineering, The Hong Kong University of Science and Technology.
 email:{\tt\small yliuhb@connect.ust.hk, zxubg@connect.ust.hk, hhuangat@connect.ust.hk, eewanglj@ust.hk, eelium@ust.hk}.
        }
}

\begin{document}

\markboth{Journal of \LaTeX\ Class Files,~Vol.~14, No.~8, August~2021}%
{Shell \MakeLowercase{\textit{et al.}}: A Sample Article Using IEEEtran.cls for IEEE Journals}

\IEEEpubid{0000--0000/00\$00.00~\copyright~2021 IEEE}

\maketitle
\thispagestyle{empty}
\pagestyle{empty}


\begin{abstract}

Predicting accurate depth with monocular images is important for low-cost robotic applications and autonomous driving. This study proposes a comprehensive self-supervised framework for accurate scale-aware depth prediction on autonomous driving scenes utilizing inter-frame poses obtained from inertial measurements. In particular, we introduce a Full-Scale depth prediction network named FSNet. FSNet contains four important improvements over existing self-supervised models: (1) a multichannel output representation for stable training of depth prediction in driving scenarios, (2) an optical-flow-based mask designed for dynamic object removal, (3) a self-distillation training strategy to augment the training process, and (4) an optimization-based post-processing algorithm in test time, fusing the results from visual odometry. With this framework, robots and vehicles with only one well-calibrated camera can collect sequences of training image frames and camera poses, and infer accurate 3D depths of the environment without extra labeling work or 3D data. Extensive experiments on the KITTI dataset, KITTI-360 dataset and the nuScenes dataset demonstrate the potential of FSNet. More visualizations are presented in \url{https://sites.google.com/view/fsnet/home}         
\end{abstract}

\def\abstractname{Note to Practitioners}
\begin{abstract}
This paper was motivated by the problem of unsupervised monocular depth for robotic deployment. We notice that PoseNet is not generalizable and by nature monodepth2 only predict depths up to a scale. We believe that we should not expect PoseNet, a ResNet on a concatenation of two images, to produce more reliable poses than the localization module in a robot.  So we try our best to completely avoid using PoseNet. This creates much unstability in training, but we managed to fix it in FSNet with multichannel output and self-distillation. We also believe the network should try to directly predict accurate depth with a correct scale at any cases. So our method could produce meaningful results on static frames or scenes with little/no VO points (same as the network's direct prediction).  There are images without VO points in our multi-frame experiment, but our method is robust enough to fix this problem.  In future research, we will include multi-frame depth predictions for more accurate depth prediction.

\end{abstract}

\begin{IEEEkeywords}
Autonomous Driving, Depth Prediction, Computer Vision.
\end{IEEEkeywords}




\section{Introduction}
\label{section:Introduction}

\begin{figure}
    \centering
    \includegraphics[width=0.5\textwidth]{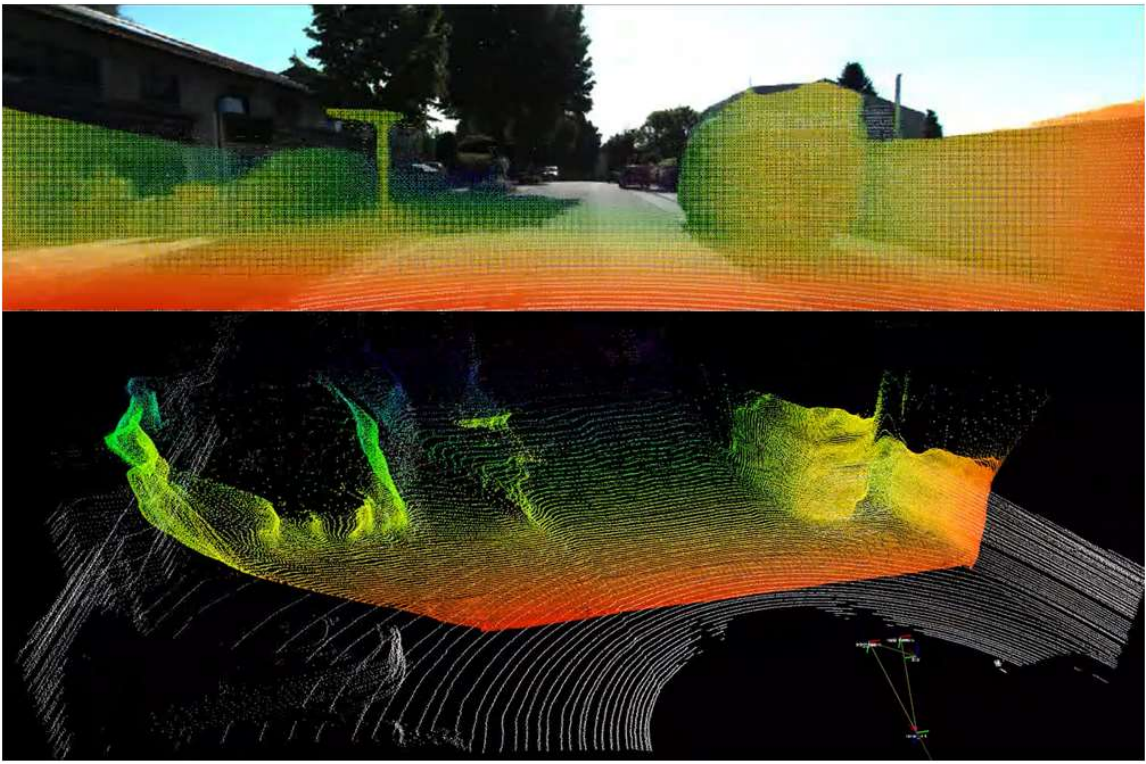}
    \caption{Prediction result sample from KITTI-360 dataset. White Points are points from LiDAR sensing. Colored points are prediction results from FSNet, which correctly express the 3D scenes with dense points.
    }
    \label{fig:kitti360_init_example}
\end{figure}

\IEEEPARstart{D}{ense} depth prediction from a single RGB image is a significant challenge in computer vision and is useful in the robotics community. Though active sensors like LiDAR can directly produce accurate depth measurements of the environment, camera-based setups are still popular because cameras are relatively cheap, power-efficient, and flexible in that they can be mounted on various robotic platforms. Thus, monocular depth prediction enables more sophisticated 3D perception and planning tasks for many camera-based robots and low-cost self-driving setups \cite{tase1,tase2,tase3}, also including products like Tesla's FSD and Valeo's vision-only system \cite{valeo_2022}. A qualitative prediction sample of our work is presented in Figure ~\ref{fig:kitti360_init_example}, showing the potential of directly perceiving the world in 3D with a monocular camera.

Efficiently deploying a monocular depth prediction network in a new environment raises challenges for existing methods. As pretrained vision models do not usually generalize well enough to new scenes or a new camera setup \cite{MonodepthSurvey2022}, it is extremely useful and convenient to directly train a depth estimation model using raw data collected from the deployed robot in the target environment. However, most of the current self-supervised monocular depth using only monocular image sequences can only provide depth prediction with an ambiguity in the global scale of the depth results unless using stereo images \cite{monodepth2, manydepth2021temporal, watson-2019-depth-hints, ramamonjisoa-2021-wavelet-monodepth} or external point clouds supervision in training \cite{FuCVPR18DORN}. 

A mobile robotic system or a self-driving car usually produces a robot's pose online by a standalone localization module. Moreover, with onboard sensors like the inertial measurement unit (IMU) and wheel encoder, the localization module can produce accurate relative poses between consecutive keyframes. Thus, we focus on the usage of poses to tackle scale ambiguity so that the network can be trained to predict depths with correct global scales.

In this paper, we develop the FSNet framework step by step from the use of pose. First, we investigate the training process of the unsupervised MonoDepth and demonstrate why directly replacing the output from PoseNet \cite{monodepth2} with real camera poses will cause failure in training. This justifies the multichannel output representation adopted in our paper. Then, the availability of poses enables the computation of optical-flow-based masks for dynamic object removal. Moreover, we introduce a self-distillation framework to help stabilize the training process and improve final prediction accuracy. Finally, to use historical poses in test time, we introduce an optimization-based post-processing algorithm that uses sparse 3D points from visual odometry (VO) to improve the test time performance and enhance the robustness to changes in the extrinsic parameters of the robot. The main contributions of the paper are as follows:
\begin{itemize}
    \item An investigation of the training process of the baseline unsupervised monocular depth prediction networks. A multichannel output representation enables stable network training with full-scale depth output.
    \item An optical-flow-based mask for dynamic object removal inspired by the introduction of inter-frame poses.
    \item A self-distillation training strategy with aligned scales to improve the model performance while not introducing additional test-time costs.
    \item An efficient post-processing algorithm that fuses the full-scale depth prediction and sparse 3D points predicted by visual odometry to improve test-time depth accuracy.
    \item A validation and ablation study of the proposed algorithms on the KITTI raw dataset \cite{Geiger2012KITTI}, KITTI-360 dataset \cite{KITTI360}, and the nuScenes dataset\cite{nuscenes2019} to test the performance of the proposed system.
\end{itemize}

The remainder of this paper is organized as follows:
Section II reviews related work. Section III introduces our FSNet. Section IV presents the experimental results and compares our framework with existing self-supervised monocular depth prediction frameworks. 
Section V presents ablation studies and discussions on the components proposed in the paper. Finally, we conclude this paper in the last section.

\section{Related Works}
\label{section:Relate}

\subsection{Self-Supervised Monocular Depth Prediction}
Monodepth2 \cite{monodepth2} sets up the baseline framework for self-supervised monocular depth prediction. The core idea of the training of Monodepth2 is to teach the network to predict dense depths that reconstruct the target image from source views. ManyDepth \cite{manydepth2021temporal} utilizes the geometry in the matching between temporal images to improve the accuracy of single image depth prediction. However, the cost volume construction slows the inference speed and makes it difficult to accelerate on robotics platforms. 

To obtain the correct scale directly from monocular depth prediction, researchers use either LiDAR \cite{Eigen2014DepthPrediction, FuCVPR18DORN, Wong2020UnsupervisedDepthVIO} or stereo cameras \cite{watson-2019-depth-hints, ramamonjisoa-2021-wavelet-monodepth, wimbauer2020monorec, yang20d3vo, ChengDepthPredictionMonoDVO} to provide supervision. 
Among stereo methods, Depth Hints \cite{watson-2019-depth-hints}, and Wavelet Depth \cite{ramamonjisoa-2021-wavelet-monodepth} use traditional stereo matching algorithms to provide direct supervision to the depth prediction network. Monorec \cite{wimbauer2020monorec} uses stereo images to provide scale for the network and to identify moving object pixels in the sequence. DNet \cite{xue2020toward} uses the ground plane estimated from the normal of the image to calibrate the global scale of the predicted depth, but it fails in the nuScenes dataset which contains many image frames without a clear ground plane.

Our proposed FSNet uses poses from robots to regularize the network to predict correct scales without requiring a multiview camera or LiDARs. We point out that relative poses between frames are commonly available for robotic platforms because of sensors like IMUs and wheel encoders.
\begin{figure*}
    \centering
    \includegraphics[width=0.9\textwidth]{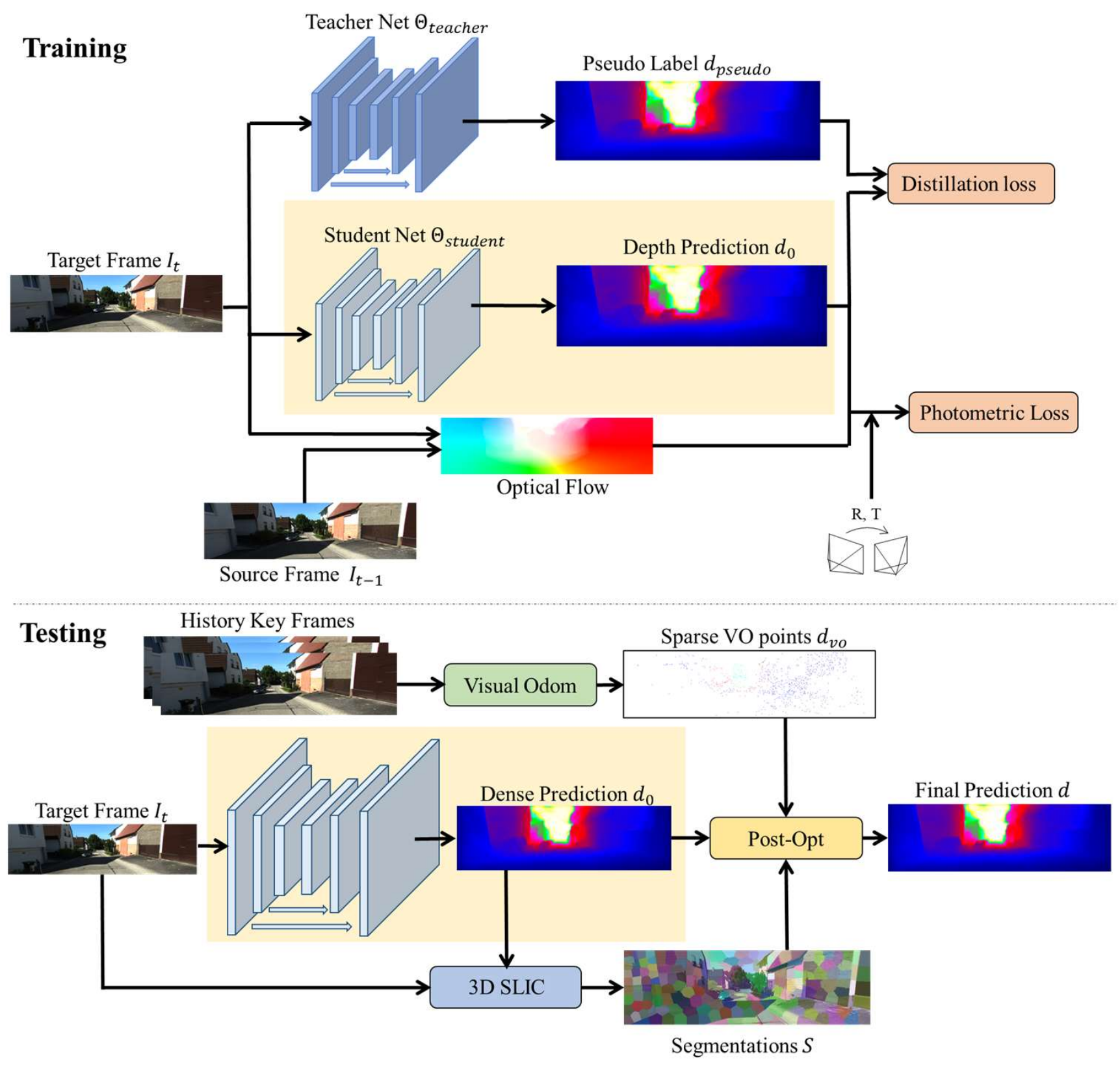}
    \caption{
    The training and inference process of FSNet. During training, the teacher network $\Theta_{teacher}$ and the student network $\Theta_{student}$ utilize a U-Net structure with a ResNet backbone.
    $\Theta_{teacher}$ is pretrained using only the photometric loss and is frozen during the training of the student network. $\Theta_{student}$ is trained with a summation of photometric loss and the distillation loss. During testing, the trained network first predict dense full-scale Depth $d_0$.  The prediction will be merged with sparse VO points $d_{vo}$ to produce accurate final prediction. The 3D SLIC algorithm is used to segment the image and limit the scale of the post-optimization problem.
    }
    \label{fig:monodepth_struct}
\end{figure*}
\subsection{Knowledge Distillation}
Knowledge distillation (KD) is a field of pioneering training methods that transfers knowledge from teacher models to target student models \cite{Hinton2015DistillingTK}. KD has been applied in various tasks, including image classification \cite{yun2020regularizing}, object detection \cite{Chen2017LearningEO, Yun2021ConflictKD}, and incremental learning \cite{yun2021KDIncre}. The philosophy of knowledge distillation is training a small target model with an additional loss function to mimic the output of a teacher model. The teacher network could be an ensemble of the target models \cite{Asif2020EnsembleKD}, a larger network \cite{Yang2020KnowledgeDV}, or a model with additional information inputs or different sensors \cite{Chong2022MonoDistillLS}.

Self-distillation (SD) is a special case where the teacher model is completely identical to the student model, and no additional input is applied to train the teacher model. Some existing works have demonstrated its performance on image classification \cite{Zhang2019BeYO}. 

In this paper, we first investigate the training process of an unsupervised monocular depth predictor, then we apply an offline-SD to our proposed model to regularize the training process and improve the final performance.

\subsection{Visual Odometry}
Visual odometry (VO) systems \cite{klein2009parallel,forster2016svo,campos2021orb} are widely used to provide robot centric pose information for autonomous navigation \cite{engel2012camera} and simultaneously estimate the ego-motion of the camera and the position of 3D landmarks. 
With IMU or Global navigation satellite system (GNSS) information, the absolute scale of the estimations can also be recovered \cite{qin2018vins,cao2022gvins}.

ORB-SLAM3 \cite{campos2021orb} is a typical indirect VO method in which the current input frame is tracked online with a 3D landmark map built incrementally.
In this tracking process, the local features extracted are first associated with landmarks on the map.
Then, the camera pose and correspondences are estimated within a Perspective-n-Point scheme.
The depth of the local feature point can be retrieved from the associated 3D landmark and the camera pose, yielding an image-aligned sparse depth map for each frame.
In this work, we use these sparse depth maps as an input for the full-scale dense depth prediction post-optimization procedure.

\subsection{Monocular Depth Completion}
Monocular depth completion means predicting a dense depth with a monocular image and sparse ground truth LiDAR points. Learning-based methods usually follow the scheme of partial differential equations (PQEs) that extrapolate depth between sparse points based on the constraints learned from RGB pixels \cite{cheng2018dcspn}.  
Other new methods without deep learning or extra ground truth labels exist. IPBasic \cite{ku2018defense} produces dense depth with morphological operations. Bryan et al. \cite{Bryan2021Deter} introduce superpixel segmentation to segment images into different units, and each unit is considered as a plane or a convex hull.

In our proposed post-processing step, the merging between dense network depth prediction and visual odometry can be formulated into the same optimization problem as monocular depth completion but with extra uncertainty and hints. The proposed method is formulated without extra LiDAR point labels.

\section{MonoDepth2 Review}

\label{sec:monodepth2}
We first review the pipeline of MonoDepth2. The target is to train a model $\mathcal{F} $ mapping the input target image frame $I_0$ to a dense depth map $d_0$, with reconstruction supervision from neighboring frames. The depth $d$ is decoded from the convolutional output $x$ as 
\begin{equation}
    \frac{1}{d} = \frac{1}{d_{max}} + \text{sigmoid}(x) * (\frac{1}{d_{min}} - \frac{1}{d_{max}}),
\end{equation}
where $d_{max}=100$ and $d_{min}=0.1$ are hyper-parameters for the boundary of the depth prediction. 

The depth map is reprojected to neighboring source keyframes with pose predicted from PoseNet.  Then, the target frame is reconstructed using colors sampled from neighboring source frames and produces $I_0^l$ where $l \in \{-1, 1\}$. The photometric loss is the weighted sum of the structural similarity index measure (SSIM) and the L1 difference between the two images. The loss is expressed as:
\begin{equation}
    l_{photo}(d_0) = \alpha \frac{1 - SSIM(I_0, I_0^l)}{2} + \beta |I_0 - I_0^l|,
\end{equation}
where $SSIM()$ measures the structural difference between the two images using a $3\times 3$ kernel, and $\alpha$ and $\beta$ is a constant with value 0.85 and 0.15, following \cite{monodepth2}.

As shown in our-own experiments and \cite{wei2022surround}, directly substituting PoseNet with poses from the localization module will cause failure in training the network. We notice that the baseline depth decoder method will produce corrupted reconstruction results at initialization if directly fed with the correct pose.  We need to take additional measures to preserve organic reconstruction results during the start of the training.

\section{Methods}
In order to stably train the depth prediction network with inter-frame poses, we reformulate the output of MonoDepth2 as multi-channel outputs. Inter-frame poses are further utilized to filter out dynamic objects during training with an optical flow-based mask. The overall system pipeline is presented in Fig~\ref{fig:monodepth_struct}.
\subsection{Output Modification From MonoDepth2}
\label{sec:modification}
\textbf{Multichannel Output:} In order to preserve organic reconstruction results, we select reformulate the output as multi-channel outputs, to allow larger depth values during initialization and also unsaturated gradients at large depth values.

Assuming the predicted depth $d$ should be within $[d_{min}, d_{max}]$ and the output is decoded from $N$ channels of the output network, then $q= \frac{d_{max}}{d_{min}}^{1/N}$ is defined as the proportion between consecutive depth bins, and the $i$th bin will represent $d_i = d_{min} \times q^{i-1}$. Considering the softmax activated output of the network being $Z \in \mathbb{R}^N$, we interpret the weighted mean of each depth bin as the predicted depth $d = \sum_i z_i \cdot d_i$. 

At initialization, the network will predict an almost-uniform distribution for each depth bin, so the initial decoded depth will be the algorithmic mean of the depth bins $d' = \frac{1}{N} \sum_i d_i$. To guide the selection of hyperparameters, we analyze the initial depth values $d'$ in our framework.

\textbf{Theorem :} The algorithmic mean $d'$ of a proportional array, bounded by $[d_{min}, d_{max}]$ will converge to the following limit as the number of bins $N$ grows: 
\begin{equation}
    \lim_{N\rightarrow\infty} d' = \frac{d_{max} - d_{min}}{\ln{d_{max}} - \ln{d_{min}}}.
\end{equation}

\textbf{Proof:} Denote the boundary of the predicted depth as $[d_{min}, d_{max}]$, the $i$ th bins  will represent $d_i = d_{min} \times q^{i-1} = d_{min} (\frac{d_{max}}{d_{min}})^{i/N}$. As presented in the paper, the initial depth predicted by the network will be the mean of all depth bins $d' =  \frac{1}{N}\sum_i{d_i}$.

The mean of the proportional array will be 

\begin{align}
\begin{split}
    \lim_{N \rightarrow \infty} d' &= \lim_{N \rightarrow \infty} \frac{1}{N}\sum_i{d_i} = 
    \lim_{N\rightarrow\infty} \frac{1}{N} \sum_i{d_{min} (\frac{d_{max}}{d_{min}})^{i/N}} \\
      &= \lim_{N \rightarrow \infty} \frac{1}{N} \frac{d_{max} - d_{min}}{1 - q} \\
      &=  (d_{max} - d_{min}) \lim_{N\rightarrow\infty} \frac{1}{N(1 - (\frac{d_{max}}{d_{min}})^{1/N})} \\
      &= (d_{max} - d_{min}) \lim_{M\rightarrow 0}\frac{M}{1 - (\frac{d_{max}}{d_{min}})^M} \\
      &\overset{\mathrm{H}}{=} \frac{d_{max}-d_{min}}{\ln{d_{max}} - \ln{d_{min}}}.
\end{split}
\end{align}

L'Hôpital's rule is applied at $\overset{\mathrm{H}}{=}$ to obtain the final result. 

We set $d_{max}=100m$ for the autonomous driving scene. Based on (4), we choose $d_{min}=0.1m$ for our network to obtain $d' > 10m$ to ensure stable training initialization.

\textbf{Camera-Aware Depth Adaption:} To train one single depth prediction network on all six cameras of the nuScenes dataset, we point out that we need to take the difference of camera intrinsic matrix into account. Similar object appearance produces similar features in the network, while depth predictions should vary based on the focal length of the camera. As a result, we adapt the value of the depth bins according to the input camera $f_x$ as $d_i = d_i^0 \cdot \frac{f_x}{f_{base}}$, where $f_{base}$ is a hyper-parameter chosen according to the front camera in nuScenes.
\subsection{Optical Flow Mask}
\label{sec:flow}
With the introduction of ego-vehicle poses at training times, we have a stronger prior for dynamic object removal. We know that the reprojection of a static 3D point on neighboring frames must stay on an epipolar line regardless of the distance between that point to the camera. The epipolar line $L$ can be computed as a vector with a length of three $L = \mathbf{F} \cdot [p_x, p_y, 1] ^T$, where $(p_x, p_y)$ are the the coordinate of the pixel in the base frame and $\mathbf{F}$ is the fundamental matrix between the two frame. The fundamental matrix $F$ is further calculated from $\mathbf{F}=K^T \cdot [t]_{x} \cdot R \cdot K$, with camera intrinsics parameters $K$, and relative pose $(R, t)$ between two image frames. 

Therefore, if the reprojected point was far from its epipolar line, we can estimate that this point is probably related to a dynamic object and should probably be omitted during training loss computation.

To construct the optical flow mask for dynamic object removal, we first compute the optical flow $(d_x, d_y)$ between the two image frames using a pretrained unsupervised optical flow estimator ArFlow \cite{liu2020arflow}. The distance $dis_l$ between the reprojected point from the epipolar line is computed as with 
\begin{equation}
    dis_l = \frac{L \cdot [d_x, d_y, 1]^T}{\sqrt{L_0^2 + L_1^2}},
\end{equation}
where $L_0, L_1$ represents the first and second element of vector $L$. Pixels with a deviation larger than 10 pixels are considered dynamic pixels. In this way, we produce a loss mask for photometric reconstruction loss computation.

\subsection{Self-Distillation}
\label{sec:distill}
As discussed in Section~\ref{sec:monodepth2}, the monocular depth prediction network starts training with a generally uniform depth map. 
The loss function, with a kernel size of $3\times 3$, only produces guaranteed optimal gradient directions to the network when the reconstruction error is within several pixels. As a result, at the start of the training, the network is first trained on pixels with ground truth depths close to the initialization state, and the gradients at other pixels will be noisy. The noise in the gradient will affect the stability of the training process. In order to stabilize the training process by increasing the signal-noise ratio (SNR) of the training gradient, we introduce a self-distillation framework. 

Using the baseline self-supervised framework, we first train an FSNet $\Theta_{teacher}$. The first FSNet will suffer from the noisy learning process mentioned above and produce sub-optimal results. Then, we train another FSNet $\Theta_{student}$ from scratch using the self-supervised framework, with pseudo label $d_{pseudo} = \Theta_{teacher}(I_0)$ from the first FSNet to guide the training. With the pseudo label from the teacher net, the student network will receive an additional meaningful training gradient from the beginning of the training phase. Thus, the problem of noisy gradients introduced above can be mitigated. The training process is summarized in the upper half of Fig. ~\ref{fig:monodepth_struct}.

However, since the pseudo label predicted from teacher FSNet is not always accurate, we encourage the student network to predict the uncertainty $\sigma$ for each pixel adaptively, alongside the depth $d_{0}$. We observe that close-up objects have smaller errors in depth prediction than far-away objects, and we encode this empirical result in our uncertainty model.  Thus, we assume that the logarithm of the depth prediction $\log(d)$ follows a Laplacian distribution with mean $\log(d_{pseudo})$ predicted by the teacher. The likelihood $p$ of the predicted depth can be formulated as:
\begin{equation}
    p(\log(d) | \log(d_{pseudo}), \sigma) = - \frac{|\log(d) - \log(d_{pseudo})|}{\sigma} - \log(\sigma).
\end{equation}
Practically, the convolutional network directly predicts the value of $\log(\sigma)$ instead of $\sigma$ to improve the numerical stability. We adopt $l_{distill} = -p(\log(d) | \log(d_{pseudo}), \sigma)$  as the training loss. 

\subsection{Optimization-Based Post Processing}
\label{sec:post}
To deploy the algorithm on a robot, we expect the system to be robust to online permutation to extrinsic parameters that could affect the accuracy of $D_{0}$ predicted by the network. We introduce a post-processing algorithm to provide an option to improve the run-time performance of the depth prediction module by merging the sparse depth map $D_{vo}$ produced from VO.

We formulate the post-processing problem as an optimization problem. 
The optimization problem over all the pixels in the image is described as:
\begin{equation}
    \begin{array}{ll}
        \underset{D_{out}}{\operatorname{minimize}} & \sum_{i} L^{d_{out}^i} \\
        \text { where } & L^{d^i} = \lambda_0 L^i_{consist} + \lambda_1^{i} L^i_{vo} \\
        & L_{consist}^i = \sum_{j} (
            \log( \frac{d^i_{0}}{d^j_{0}}) -
            \log( \frac{d_{out}^i}{d_{out}^j}))^2 \\
        & L_{vo}^i = [\log(d^i_{vo} / d_{out}^i)] ^ 2
    \end{array}.
\end{equation}

The consistency term $L_{consist}$ is defined by the change in relative logarithm between each pixel compared to $D_{0}$. The visual odometry term $L_{vo}$ works only at a subset of points with sparse depth points, and $\lambda_1^i$ is zero for pixels without VO points.

The problem above is a convex optimization problem that can be solved in polynomial time. However, the number of pixels is too large for us to solve the optimization problem at run time. 


Therefore, we downscale the problem by segmenting images with super-pixels and simplifying the computation inside each super-pixel. The full test-time pipeline is summarized in the second half of Fig. ~\ref{fig:monodepth_struct}.

\subsubsection{3D SuperPixel Segmentation}
The segmentation method we proposed is based on Simple linear iterative clustering (SLIC) which is similar to a K-means algorithm operated on the LAB image. We propose to utilize the full-scale dense depth prediction from the network, including the difference in absolute depth $\Delta_{dep}$ to enhance the distance metric. 

\begin{algorithm}
    \caption{3D SLIC with Depth}\label{alg:cap}
    \textbf{Input} LAB image $I_{lab}$, depth image $D_{0}$.  \\
    \textbf{Output} Set of point sets $S$,\\
    \textbf{Parameters} Grid step $s$, \\cost weights $\Lambda_{slic}=\{\lambda_{lab}, \lambda_{d}, \lambda_{pix}\}$, max iteration $E$
    \begin{algorithmic}[1]
    \State Initialize a grid of cluster centers on the image coordinate. $f^k=\{I_{lab}^k, \vec{X^k}, d^k\}$ with the step size $s$.
    \For {$iteration=1,2,\ldots, E$}
		\For {each pixel $i$}
		    \State Compute the distance to cluster centers $l_{ik} = L_{slic3d}(f^k, f^i, \Lambda)$
		    \State Assign to the closest center set $S^k$.
		\EndFor
		
		\For {each center $k$}
		    \State Update the center vector with the mean among point sets $S^k$
		\EndFor
    \EndFor
    \end{algorithmic}
\end{algorithm}

The feature of each pixel is composed of the LAB color channel $I_{lab}^i$, the coordinate in image frame $X^i$ and the depth $d^i$. The distance between each pixel and the cluster center is the weighted sum of the three distances:

\begin{align*}
    L_{slic3d}(f^k, f^i, \Lambda) &=  \lambda_{lab} \cdot norm(I^k_{lab} - I^i_{lab})  \\
     &+ \lambda_{d} \cdot |d^k - d^i| \\
     &+ \lambda_{pix} \cdot norm(X^k, X^i)
\end{align*}

The proposed 3D SLIC algorithm is presented in Algorithm ~\ref{alg:cap}. The output will be a set of point sets $S_k$. Notice that we utilize a GPU to accelerate the process.

\subsubsection{Optimization Reformulation}

After obtaining the segmentation results from 3D SLIC, we re-formulate the problem as a nested optimization problem to simplify the computation. The inner optimization computes scale changes for each pixel while the outer optimization considers the constraints between different segments.

\textbf{Inner Optimization:} For each point cluster, we assume it describes a certain geometric unit and the dense depth is correct up to a scale. The inner-optimization target is determining a scale factor $v$ so that the difference between sparse VO points and the predicted depth is minimized. The optimization for points in each cluster can be formulated as:

\begin{equation}
    \begin{array}{ll}
        \underset{\log{v}}{\operatorname{minimize}} & \frac{1}{2} \sum_i^{N_{vo}^k}(\log{d^i_0} + \log{v} - \log{d_{vo}^i})^2 \\
    \end{array}.
\end{equation}

The solution to the inner optimization problem can be derived as:

\begin{align}
   \frac{\partial L}{\partial \log{v}} &= \sum_i^{N_{vo}^k} (\log{d^i_0} + \log{v} - \log{d_{vo}^i}) = 0 \\ 
   \log{v} &= \frac{1}{N_{vo}^k} \sum_i^{N_{vo}^k}\log\frac{d^i_{vo}}{d^i} = lg^k_{vo}.
\end{align}

\textbf{Outer Optimization:} Rewriting the original pixel-wise optimization problem into a segment-wise one forms:

\begin{equation}
    \begin{array}{ll}
        \underset{lg^k}{\operatorname{minimize}} & L = \sum_{k} L^{lg^k} \\
        \text { where } & L^{lg^k} = \lambda_0 L^{lg^k}_{consist} +  \lambda_1^{k} L^i_{vo} +
        \lambda_2(lg^k - lg^k_0)^2\\
        & L_{consist}^i = \sum_{j} [
            (lg^k - lg^j) - (lg^k_0 - lg^j_0)
            ]^2 \\
        & L_{vo}^i = (lg^k_{tar} - lg^k) ^ 2,
    \end{array}
\end{equation}
where $lg^k_{tar}$ indicates the target for each cluster computed by inner optimization, and $\lambda_1^k=\lambda_1$ if there are VO points inside the $k$-th segment and $0$ otherwise. 

For this convex optimization problem, a solution that satisfies the Karush–Kuhn–Tucker (KKT) condition will be the globally optimal solution. The KKT condition implies the differential of $L$ w.r.t. each variable $lg^k$ being zero:
\begin{align*}
   [(N-1) \lambda_0 + \lambda_1^{k} + \lambda_2] lg^k - \lambda_0 \sum_{i\neq k} lg^i &=\\
   \lambda_2 lg^k_0  + \lambda_1^{k} lg^k_{tar} + &\lambda_0 \sum_{i\neq k} (lg^k_0 - lg^i_0),
\end{align*}
which forms $N$ linear equations. We denote 
$$\Lambda = (N-1) \lambda_0 + \lambda_1^{k} + \lambda_2.$$
The solution vector $\vec{lg}=[lg^0, lg^1,...]^T$ to the system of the linear equation will be $\vec{lg} = A^{-1}B$, where:

\begin{align}
    A &= \begin{bmatrix}
        \Lambda & - \lambda_0  & \cdots & -\lambda_0 \\
        - \lambda_0 & \Lambda  & \cdots & -\lambda_0\\
        \vdots & \vdots  & \ddots   & \vdots  \\
        -\lambda_0 & -\lambda_0  & \cdots\  & \Lambda  \\
        \end{bmatrix} \\
    B &= [\cdots, \lambda_2 lg^k_0  + \lambda_1^{k} lg^k_{tar} + \lambda_0 \sum_{i\neq k} (lg^k_0 - lg^i_0), \cdots]^T.
\end{align}

The overall post-optimization algorithm is presented at Algorithm~\ref{alg:post}. 
\begin{algorithm}
    \caption{Post-Optimization}\label{alg:post}
    \textbf{Input} Log-depth image $lg_{net}$, VO depth image $D_{vo}$,  point sets $S$\\
    \textbf{Output} Optimized log-Depth $lg$,
    \begin{algorithmic}[1]
    \For {each cluster $S^k$}
        \State Compute mean-log-depth $lg_0^k = \frac{1}{N^k}\sum_i(lg_{net}^i)$.
        \State Compute optimized $v$ from Equation (14).
        \State Obtain target $lg_{tar}^k = lg_0^k \cdot v$.
    \EndFor
    \State Compute the optimized center from outer optimization $\vec{lg}_{seg} = A^{-1}B$
    \For {each cluster $S^k$}
		\For {each pixel $i$ in the cluster}
		    \State Obtain final log-depth $lg^i = lg_{net}^i \cdot \frac{lg_{seg}^k}{lg_0^k} $
		\EndFor
    \EndFor
    \end{algorithmic}
\end{algorithm}

The solution involves the inverse of an $N\times N$ matrix $A$, whose computational complexity scale grows in $\mathcal{O}(n^3)$. This explains the necessity of downscaling the pixel-wise optimization problem into a segment-wise problem with the proposed 3D SLIC algorithm and the use of a nested optimization scheme.

\begin{table*}[]
\caption{Performance of full-scale MonoDepth on KITTI and nuScenes.
Results on nuScenes are averaged over six cameras. For scale factor, "GT" is using LiDAR median scaling methods and "None" means we directly evaluate the error without post-processing scale. "\textbf{*}" indicates methods using sequential images in test time. The \cellcolor{red!25}{pink} columns are error metrics, the lower the better; the \cellcolor{blue!25}{blue} columns are accuracy metrics, the higher the better.}
    \label{tab:nusc_test}
    \centering
    \def\arraystretch{1.3}
    \begin{tabular*}{0.89\textwidth}{|l|l|c|c|c|c|c|c|c|c|}
    \cline{1-10} {\bf Data}&{\bf Methods} & {\bf Scale Fac.} &  \cellcolor{red!25}{Abs Rel } & \cellcolor{red!25}{Sq Rel} & \cellcolor{red!25}{\bf RMSE} & \cellcolor{red!25}{RMSE LOG} & \cellcolor{blue!25}{$\delta<1.25$} & \cellcolor{blue!25}{$\delta<1.25^2$} & \cellcolor{blue!25}{$\delta<1.25^3$} \\ 
    \cline{1-10}
    \multirow{9}{*}{KITTI}& Bian et al. \cite{Bian2019Depth}&\multirow{6}{*}{GT} & 0.128 & 1.047 & 5.234 & 0.208 & 0.846 & 0.947 & 0.976 \\
    & CC \cite{cc2019ranjan}& & 0.139 & 1.032 & 5.199 & 0.213 & 0.827 & 0.943 & 0.977 \\
    & MonoDepth \cite{monodepth2} &   &   0.116  &   0.903  &   4.863  &   0.193  &   0.877  &   0.959  &   0.981  \\
    & DNet \cite{xue2020toward} &  &  0.113  &   0.864  &   4.812  &   0.191  &   0.877  &   0.960  &   0.981  \\
    &\textbf{FSNet (single frame)} &   &   \textbf{0.113}  &   \textbf{0.857}  &   \textbf{4.623}  &   \textbf{0.189}  &   \textbf{0.876} &   \textbf{0.960}  &   \textbf{0.982}  \\
    &  \textbf{*FSNet (post-opt)}  &   & \textbf{0.111}  &   \textbf{0.829}  &   \textbf{4.393}  &   \textbf{0.185}  &   \textbf{0.883}  &   \textbf{0.961}  &   \textbf{0.983}  \\
    \cline{2-10}
    & DNet \cite{xue2020toward} &  &  0.118  &   0.925  &   4.918  &   0.199  &   0.862  &   0.953  &   0.979 \\
    &   \textbf{FSNet (single frame)} &  \multirow{2}{*}{None}   & \textbf{0.116}  &   \textbf{0.923}  &   \textbf{4.694}  &   \textbf{0.194}  &   \textbf{0.871}  &   \textbf{0.958}  &   \textbf{0.981}  \\
    &   \textbf{*FSNet (post-opt)}   &   & \textbf{0.109}  &   \textbf{0.866}  &   \textbf{4.450}  &   \textbf{0.189}  &   \textbf{0.879}  &   \textbf{0.959}  &   \textbf{0.982}  \\
    
    \multirow{5}{*}{Nusc}& MonoDepth \cite{monodepth2} & \multirow{3}{*}{GT}  &   0.233  &   4.144  &   6.979  &   0.308  &   0.782  &   0.901  &   0.943  \\
    &\textbf{FSNet (single frame)} &  &   \textbf{0.239}  &   \textbf{5.104}  &   \textbf{6.979}  &   \textbf{0.308}  &   \textbf{0.794}  &   \textbf{0.904}  &   \textbf{0.942}  \\
    &\textbf{FSNet (multiframe)} &  &   \textbf{0.235}  &   \textbf{4.503}  &   \textbf{6.923}  &   \textbf{0.307}  &   \textbf{0.786}  &   \textbf{0.895}  &   \textbf{0.937}  \\
    \cline{2-10}
    &\textbf{FSNet (single frame)} & \multirow{2}{*}{None} &   \textbf{0.238}  &   \textbf{6.180}  &   \textbf{6.865}  &   \textbf{0.319}  &   \textbf{0.806}  &   \textbf{0.904}  &   \textbf{0.940}  \\
    &\textbf{FSNet (multiframe)} &  &   \textbf{0.238}  &   \textbf{6.198}  &   \textbf{6.489}  &   \textbf{0.311}  &   \textbf{0.811}  &   \textbf{0.910}  &   \textbf{0.944}  \\
    \cline{1-10}
    \end{tabular*}
    
\end{table*}

\section{Experiments}

\subsection{Experiment Settings}
We first present the dataset and background settings of our experiments.

We utilize the following datasets in our experiments to evaluate the performance of our approach:

\begin{itemize}
    \item KITTI Raw dataset \cite{Geiger2012KITTI}: This dataset was designed for autonomous driving and many existing works have produced official results on it. We mainly evaluate FSNet on the Eigen Split \cite{Eigen2014DepthPrediction}. It contains 39810 monocular frames for training and 697 images from multiple sequences for testing. Images are sub-sample to $192\times 640$ during training and inference for the network.
    \item NuScenes dataset \cite{nuscenes2019}: This dataset contains 850 sequences collected with six cameras around the ego-vehicle.  We separate the dataset under the official setting with 700 training sequences and 150 validation sequences. We only select scenes without rain and night scenes during both training and validation.  We uniformly sub-sample the validation set for validation.  Images are sub-sampled to $448 \times 762$. Unlike \cite{Vitor2022FSM}, we treat images at each frame as six independent samples during both training and testing. The model will have to adapt to different camera intrinsic and extrinsic parameters. Furthermore, because the lidar and the cameras are not synchronized, there will be noise in the poses between frames. FSNet needs to overcome these problems to achieve stable training, and also predict scale-aware depths.
\end{itemize}

\textbf{Depth Metrics:} Previous works, including \cite{monodepth2, manydepth2021temporal, wimbauer2020monorec, watson-2019-depth-hints} mainly focus on metrics where the depth prediction is first aligned with the ground truth point clouds using a global median scale before computing errors. In this paper, we first present data on the scaled metrics for comparison with existing methods. Then, we focus on metrics \textbf{without} median scaling in ablation studies.

\textbf{Data Augmentation:} Besides photometric augmentation adopted in MonoDepth2 \cite{monodepth2}, we implement horizontal flip augmentation for image-pose sequences, where we also horizontally flip the relative poses between image frames.

\textbf{FSNet Setting:} We adopt ResNet-18 \cite{He2015Resnet} as the backbone encoder for KITTI datasets following prior mainstream works for a fair comparison, and we adopt ResNet-34 for the nuScenes dataset because of the increasing difficulty. The PoseNet is dropped and poses from the dataset are directly used in image reconstruction, and no additional modules are created except for a frozen teacher net during distillation. In summary, we basically share the inference structure of MonoDepth2, and we do not train a standalone PoseNet. In the KITTI dataset, FSNet spends 0.03s on network inferencing and 0.04s on post-optimization, measured on RTX 2080Ti. We point out that multichannel output does not noticeably increase network inference time with additional width only in the final convolution layer.

\subsection{Single Camera Prediction}
The performance on the KITTI dataset is presented in Table~\ref{tab:nusc_test}. We point out that, even though FSNet spends extra network capability to memorize the scale of the objects in the scenes, it can produce even better depth maps than baseline models by utilizing multi-channel output, flow mask, and self-distillation.

We present results on both single-frame settings and multi-frame settings. We could appreciate the improvement from post-optimization in the result table. We further point out that our method  decouples the prediction of a single frame and post-optimization into standalone modules, which makes it flexible for different application settings. 

\begin{figure*}
    \centering
    \includegraphics[width=1.0\textwidth]{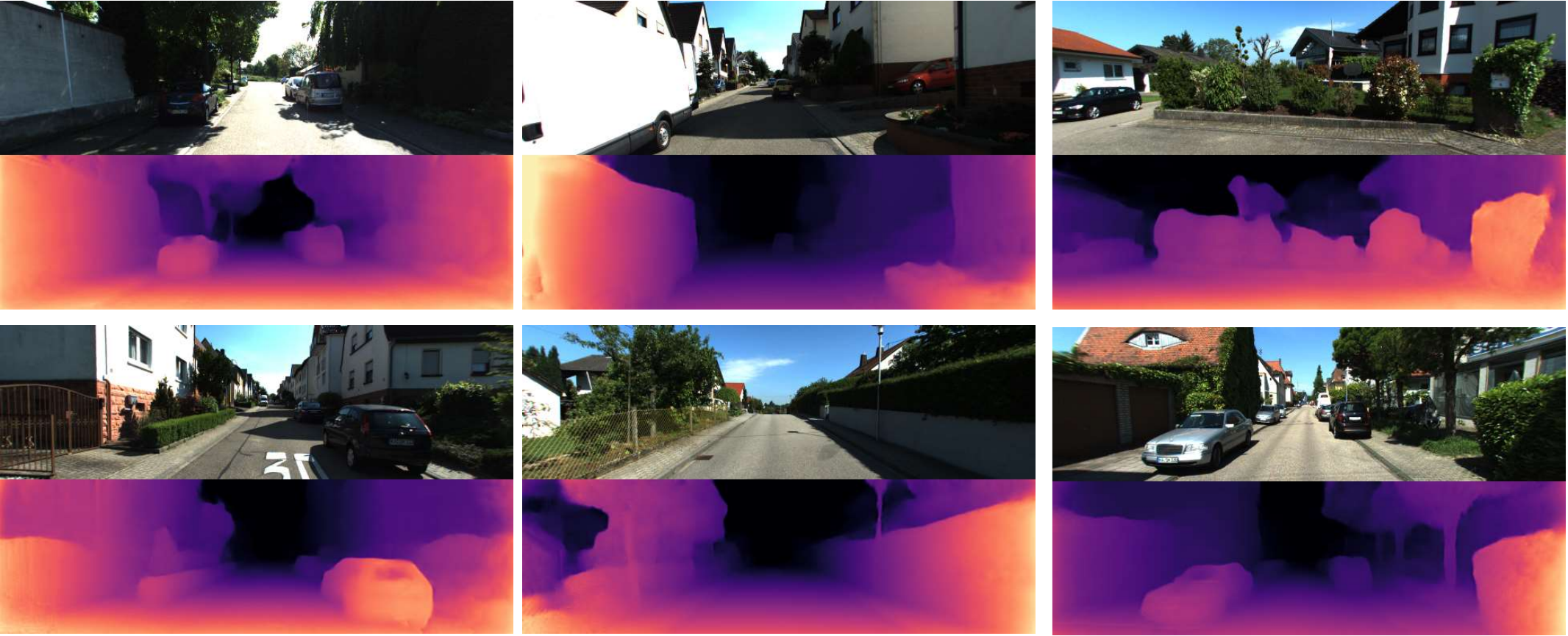}
    \caption{Prediction result sample from KITTI-360 dataset. Pixels colored in white are expected to be closer to the camera. The pictures demonstrates the network's ability to distinguish close-up objects against backgrounds. 
    }
    \label{fig:kitti360_frame_examples}
\end{figure*}

\begin{figure*}
    \centering
    \includegraphics[width=0.9\textwidth]{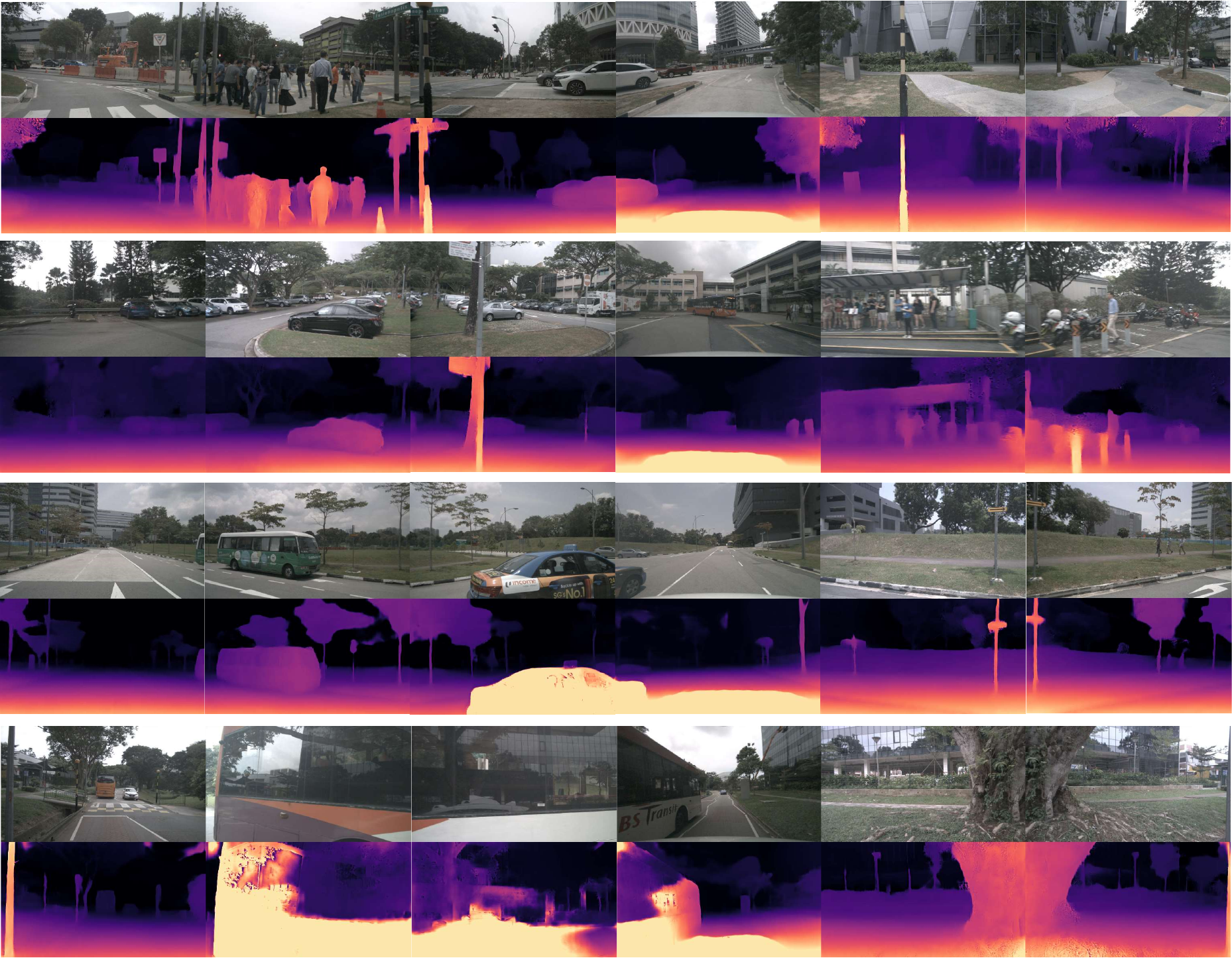}
    \caption{Self-Supervised Depth Estimation FSNet result on the nuScenes dataset. Complicated scenarios, varying camera parameters, close-up objects and large-scale specular reflection pixels make nuScenes a particularly challenging dataset.
    }
    \label{fig:nusc_vis}
\end{figure*}

\subsection{Multi-Camera Depth Prediction}

The performance on the nuScenes dataset is presented in Table ~\ref{tab:nusc_test}. We also present the detailed performance result of RMSE and RMSE log on all six cameras on the nuScenes dataset.
Some other methods \cite{Vitor2022FSM, wei2022surround} train with all six camera images in a frame at once, and propagate information between images during training and inference. We treat the data as six \textbf{independent} monocular image streams to train and infer with a single FSNet model. The single FSNet model aligns its predicted depth with different cameras using their intrinsic camera parameter. 

Because the cameras and LiDAR data are not synchronized, the relative poses between frames in the nuScenes dataset are not as accurate as those in the KITTI dataset. However,  experiment data show that FSNet trained with noisy poses can still produce depth predictions with the correct scale as the 3D scenes  and it is robust to different camera setups.

Finally, we present some qualitative results on the validation split of the nuScenes dataset in Figure~\ref{fig:nusc_vis}. FSNet predicts depth in each image independently, and we concatenate the predictions together to form the results here. The first three rows demonstrate the network's ability to identify different objects of interest in urban road scenes. The final row presents a failure case where a bus with a huge reflective glass dominates the image. More 3D visualization is presented at the project page \url{https://sites.google.com/view/fsnet/home}, which can further show that the depths predicted from the six cameras are consistent, and we can obtain a detailed 3D perception result of the surrounding environment.

\begin{table*}[]
\caption{ RMSE and RMSE Log of full-scale MonoDepth on nuScenes on all six cameras. For scale factor, "GT" is using LiDAR median scaling methods and "None" means we directly evaluate the error without post-processing scale. "\textbf{*}" indicates methods using sequential images in test time. }
    \label{tab:nusc_test_detailed}
    \centering
    \def\arraystretch{1.3}
    \begin{tabular*}{1.0\textwidth}{|l|l|cc|cc|cc|cc|cc|cc|cc|}
\cline{1-16}
                                   &                              & \multicolumn{2}{c|}{Front}                                                      & \multicolumn{2}{c|}{F.Left}                                                     & \multicolumn{2}{c|}{F.Right}                                                    & \multicolumn{2}{c|}{B.left}                                                     & \multicolumn{2}{c|}{B.right}                                                    & \multicolumn{2}{c|}{Back}                                                       & \multicolumn{2}{c|}{Avg}                                                        \\ \cline{3-16} 
\multirow{-2}{*}{\textbf{Methods}} & \multirow{-2}{*}{Fac.} & \multicolumn{1}{c|}{\cellcolor[HTML]{FFCCC9}rmse} & \cellcolor[HTML]{FFCCC9}log & \multicolumn{1}{c|}{\cellcolor[HTML]{FFCCC9}rmse} & \cellcolor[HTML]{FFCCC9}log & \multicolumn{1}{c|}{\cellcolor[HTML]{FFCCC9}rmse} & \cellcolor[HTML]{FFCCC9}log & \multicolumn{1}{c|}{\cellcolor[HTML]{FFCCC9}rmse} & \cellcolor[HTML]{FFCCC9}log & \multicolumn{1}{c|}{\cellcolor[HTML]{FFCCC9}rmse} & \cellcolor[HTML]{FFCCC9}log & \multicolumn{1}{c|}{\cellcolor[HTML]{FFCCC9}rmse} & \cellcolor[HTML]{FFCCC9}log & \multicolumn{1}{c|}{\cellcolor[HTML]{FFCCC9}rmse} & \cellcolor[HTML]{FFCCC9}log \\\cline{1-16}
MonoDepth\cite{monodepth2}                          &                              & \multicolumn{1}{c|}{6.992}                        & 0.222                       & \multicolumn{1}{c|}{6.905}                        & 0.319                       & \multicolumn{1}{c|}{7.655}                        & 0.359                       & \multicolumn{1}{c|}{5.996}                        & 0.305                       & \multicolumn{1}{c|}{7.606}                        & 0.375                       & \multicolumn{1}{c|}{6.718}                        & 0.267                       & \multicolumn{1}{c|}{6.979}                        & 0.308                       \\
\textbf{FSNet}                     &                              & \multicolumn{1}{c|}{6.999}                        & 0.225                       & \multicolumn{1}{c|}{6.652}                        & 0.312                       & \multicolumn{1}{c|}{7.703}                        & 0.362                       & \multicolumn{1}{c|}{6.137}                        & 0.311                       & \multicolumn{1}{c|}{7.047}                        & 0.362                       & \multicolumn{1}{c|}{7.265}                        & 0.275                       & \multicolumn{1}{c|}{\textbf{6.930}}               & \textbf{0.306}              \\ 
\textbf{*FSNet-Opt}                & \multirow{-3}{*}{GT}         & \multicolumn{1}{c|}{6.918}                        & 0.223                       & \multicolumn{1}{c|}{6.866}                        & 0.318                       & \multicolumn{1}{c|}{7.586}                        & 0.357                       & \multicolumn{1}{c|}{5.963}                        & 0.304                       & \multicolumn{1}{c|}{7.562}                        & 0.373                       & \multicolumn{1}{c|}{6.644}                        & 0.264                       & \multicolumn{1}{c|}{\textbf{6.923}}               & \textbf{0.307}              \\\cline{1-16}
\textbf{FSNet}                     &                              & \multicolumn{1}{c|}{6.658}                        & 0.223                       & \multicolumn{1}{c|}{6.632}                        & 0.322                       & \multicolumn{1}{c|}{7.853}                        & 0.364                       & \multicolumn{1}{c|}{5.681}                        & 0.303                       & \multicolumn{1}{c|}{7.658}                        & 0.377                       & \multicolumn{1}{c|}{6.706}                        & 0.329                       & \multicolumn{1}{c|}{\textbf{6.865}}               & \textbf{0.319}              \\ 
\textbf{*FSNet-Opt}                & \multirow{-2}{*}{None}       & \multicolumn{1}{c|}{6.902}                        & 0.235                       & \multicolumn{1}{c|}{6.396}                        & 0.323                       & \multicolumn{1}{c|}{7.317}                        & 0.352                       & \multicolumn{1}{c|}{4.965}                        & 0.282                       & \multicolumn{1}{c|}{6.901}                        & 0.361                       & \multicolumn{1}{c|}{6.584}                        & 0.276                       & \multicolumn{1}{c|}{\textbf{6.489}}               & \textbf{0.311}              \\\cline{1-16}
\end{tabular*} 
    
\end{table*}

\section{Discussions and Ablation Study}
\begin{table}
    \centering
    \caption{Abalation study of single frame prediction in FSNet on KITTI Eigen Split without scale factor.}
    \label{tab:kitti_ablation_single}
    \def\arraystretch{1.3}
    \begin{tabular*}{0.5\textwidth}{ |l|c|c|c|c|}
        \hline
        {\bf Variants} & \cellcolor{red!25}{Abs Rel$\downarrow$} & \cellcolor{red!25}{Sq Rel$\downarrow$} & \cellcolor{red!25}{\bf RMSE$\downarrow$} & \cellcolor{red!25}{RMSE log$\downarrow$} \\ 
        \hline
        biased MonoDepth &   0.126  &   1.033  &   5.151  &   0.201  \\
        exp MonoDepth &   0.138  &   1.122  &   5.479  &   0.220  \\
        \hline
        MultiChannel &   0.117  &   0.936  &   4.910  &   0.202   \\
        + Flow Mask &   0.118  &   0.928  &   4.861  &   0.200\\
        + Distillation & 0.116  &   0.923  &   4.694  &   0.194\\
        \hline
    \end{tabular*}
\end{table}

In this section, we start by focusing on how each proposed component improves performance in a single-frame setting. Then, we study the parameters and decision choices in the optimization-based post-processing.

\subsection{Single-Frame Evaluation}
In Section~\ref{sec:modification}, we introduce a multi-channel output to allow organic image reconstruction at initialization to boost-trap the training. Here, we experiment with two other output settings: (1) original monodepth2 output with a bias in the output layer; (2) exponential activation. As presented in Table~\ref{tab:kitti_ablation_single}, multi-channel output performs better at most error metrics. The original output representation of monodepth2 saturates in a common depth range like $d > 10m$, which makes it difficult to train the network. We highlight that the un-scale baseline MonoDepth tends to maintain the output in range with sufficient gradients by adjusting the scale of the pose prediction. The exponential activation, though effective in monocular 3D object detection \cite{Chen2020MonoPair},  results in un-controllable activation growth in background pixels like the sky, where the correct depths are essentially infinity, corrupting the training gradients. The experiments and analysis show that the proposed multi-channel output enables stable training and produces accurate predictions.

We further present the results with a flow mask and self-distillation in Table~\ref{tab:kitti_ablation_single}. Each proposed method incrementally improves the depth estimation results. Squared Relative (Sq Rel) and Root Mean Square Error (RMSE) improve the most, which means that the proposed methods mostly prevent the network from making significant mistakes like with dynamic objects.
We note that both methods do not introduce extra cost in the inference time, but it regulates the training process to improve performance.

\subsection{Post-Optimization Evaluation}

There are many factors contributing to the performance of the post-optimization step. This section investigates two factors: (1) the usage of depth in the 3D SLIC algorithm and (2) the importance of each loss weight in the optimization step.

The results are presented in Table~\ref{tab:kitti_ablation}. The proposed 3D SLIC algorithm improves the segmentation quality by utilizing the depth predicted by the network, thus improving the post-optimization results. When $\lambda_0=0$, the depth prediction of each super segment will be independent, and the optimization from visual odometry cannot fully propagate throughout the map. However, when $\lambda_2=0$, we completely ignore the scale predicted by the network, and errors in visual odometry, especially at dynamic objects, will corrupt the prediction result.  

\begin{table}
    \centering
    \caption{Abalation study of post-processing in FSNet on KITTI Eigen Split without scale factor.}
    \label{tab:kitti_ablation}
    \def\arraystretch{1.3}
    \begin{tabular*}{0.465\textwidth}{ |l|c|c|c|c|}
        \cline{1-5}
        {\bf Methods} & \cellcolor{red!25}{Abs Rel$\downarrow$} & \cellcolor{red!25}{Sq Rel$\downarrow$} & \cellcolor{red!25}{\bf RMSE$\downarrow$} & \cellcolor{red!25}{RMSE log$\downarrow$}  \\ 
        \cline{1-5}
        FSNet(post-opt) &   0.109  &   0.866  &   4.450  &   0.189  \\
        \cline{1-5}
        w 2D SLIC &   0.110  &   0.881  &   4.495  &   0.190   \\
        w $\lambda_0=0$ &   0.111  &   0.924  &   4.513  &   0.194\\
        w $\lambda_2=0$ & 0.121  &   0.977  &   4.585  &   0.199\\
        \cline{1-5}
    \end{tabular*}
\end{table}

\section{Conclusion}

This paper proposed FSNet, a redesign of the self-supervising monocular depth prediction a FULL-Scale self-supervised monocular depth prediction method suitable for robotic applications. First, we conducted experiments to improve our understanding of how an unsupervised MonoDepth prediction network trains and introduced our multi-channel output representation for stable initial training. Then, we developed an optical-flow-based dynamic object removal mask and a self-distillation scheme for high-performance training. Finally, we introduced a post-processing method that improves our algorithm's test-time performance with sparse 3D points produced from visual odometry. We conducted extensive experiments to validate the results of our proposed method. 

We emphasize that our proposed method requires only sequences of images and corresponding poses to train, which is straightforward for a calibrated robot with localization modules. We were able to obtain 3D information on the environment at test time. The test-time model is also efficient, and little extra cost is required. The modular design of the proposed method will also benefit the management of the actual development of an FSNet module.




\bibliographystyle{IEEEtran}

\bibliography{reference}


\begin{IEEEbiography}[{\includegraphics[width=1in,height=1.25in,clip,keepaspectratio]{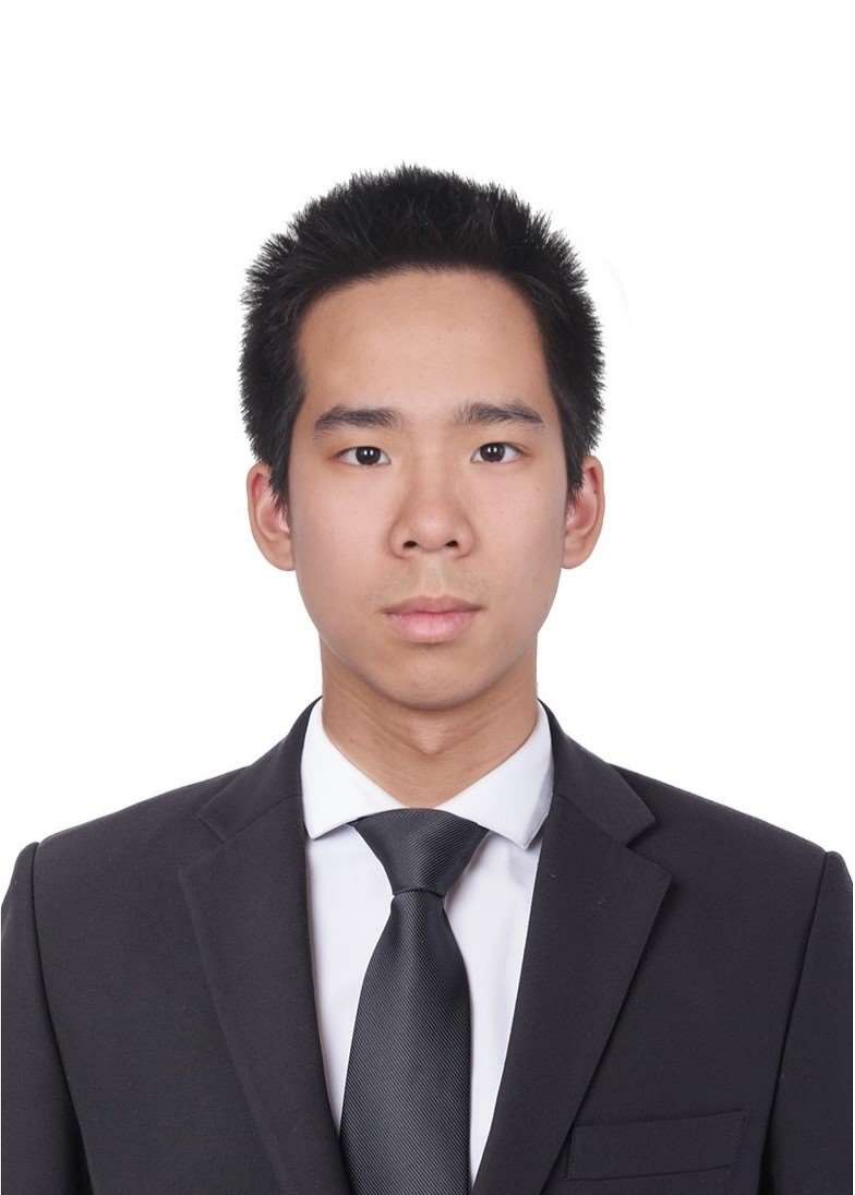}}]{Yuxuan Liu} (Student Member 2022) received his Bachelor’s degree from Zhejiang University, Zhejiang, China in 2019, majoring in Mechatronic. He is now a Ph.D candidate at the Department of Electronic and Computer Engineering, The Hong Kong University of Science and Technology, Hong Kong, China. His current research interests include autonomous driving, deep learning, robotics, visual 3D object detection, visual depth prediction, etc.
\end{IEEEbiography}
\vspace{11pt}

\begin{IEEEbiography}[{\includegraphics[width=1in,height=1.25in,clip,keepaspectratio]{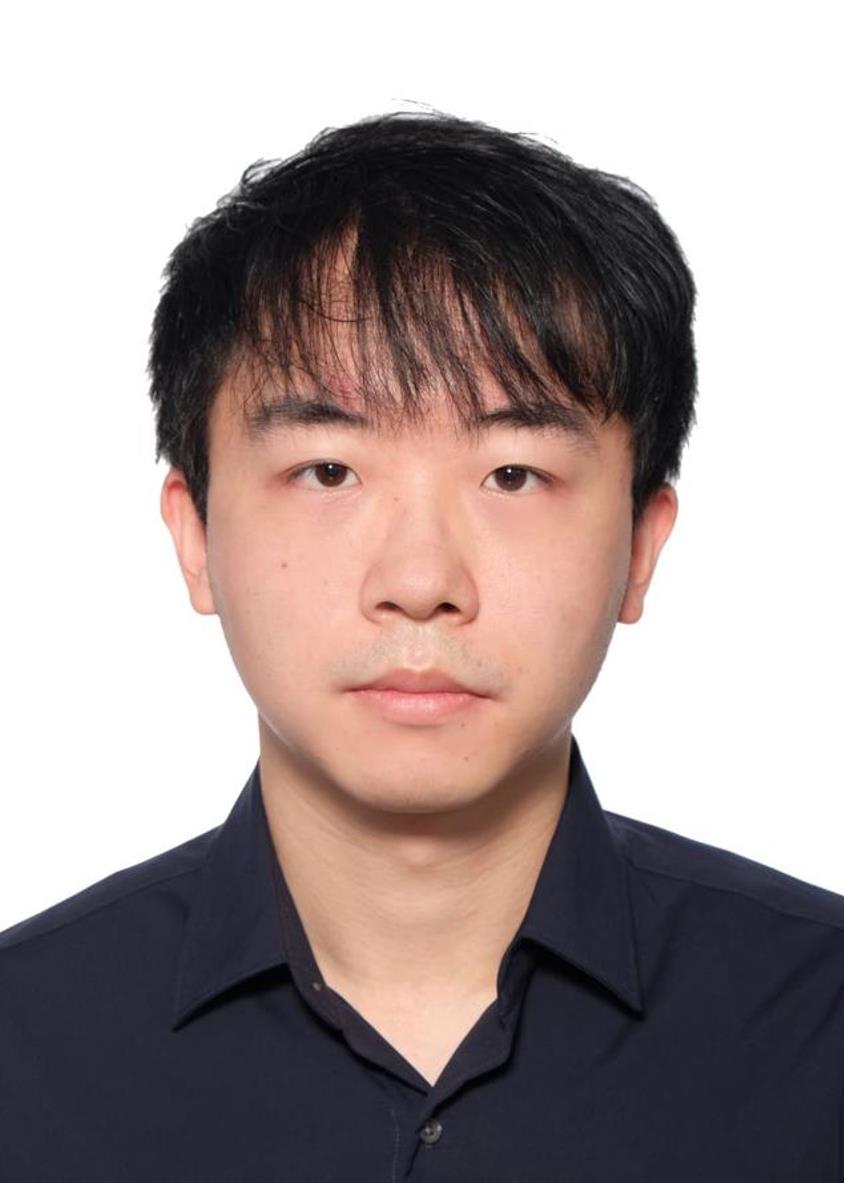}}]{Zhenhua Xu} (Student Member 2022)  received the bachelor’s degree from Harbin Institute of Technology, Harbin, China, in 2018. He is now a PhD candidate supervised by Prof. Ming Liu and Prof. Huamin Qu at the Department of Computer Science and Engineering, The Hong Kong University of Science and Technology, HKSAR, China. His current research interests include HD map automatic annotation, line-shaped object detection, imitation learning, autonomous driving, etc.
\end{IEEEbiography}
\vspace{11pt}

\begin{IEEEbiography}[{\includegraphics[width=1in,height=1.25in,clip,keepaspectratio]{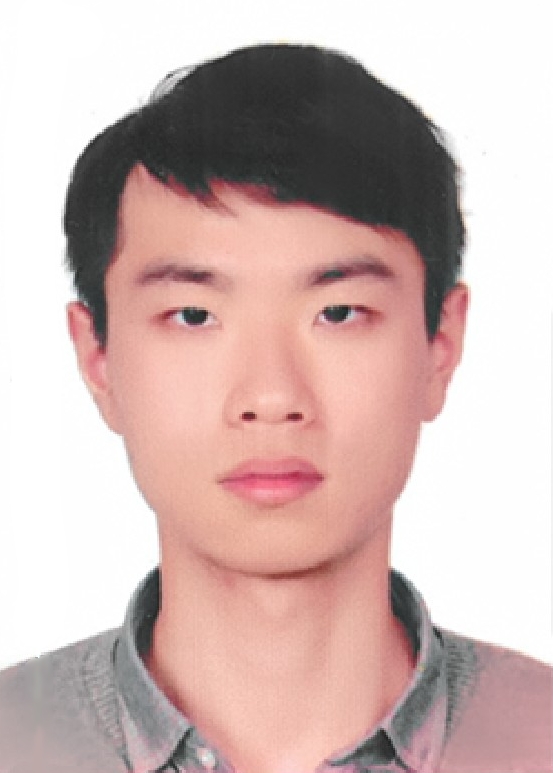}}]{Huaiyang Huang} (Student Member 2022) received the B.Eng. degree in mechatronics from Zhejiang University, Hangzhou, China, in 2018. He is now pursuing Ph.D. degree at the Department of Computer Science and Engineering, the Hong Kong University of Science and Technology, Hong Kong. His current research interests include state estimation for robotics, visual localization and visual navigation.

\end{IEEEbiography}
\vspace{11pt}

\begin{IEEEbiography}[{\includegraphics[width=1in,height=1.25in,clip,keepaspectratio]{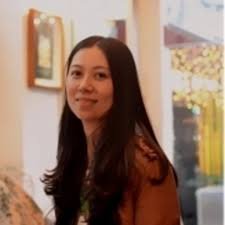}}]{Lujia Wang}  (Member 2022) received the Ph.D. degree from the Department of Electronic Engineering, The Chinese University of Hong Kong, Hong Kong, in 2015. She was a Research Fellow with the School of Electrical Electronic Engineering, Nanyang Technological University, Singapore, from 2015 to 2016. She was an associate professor with the Shenzhen Institutes of Advanced Technology, Chinese Academy of Sciences, Shenzhen, Guangdong, from 2016-2021. Her current research interests include Cloud Robotics, Lifelong Federated Robotic Learning, Resource/Task Allocation for Robotic Systems, and Applications on Autonomous Driving.
\end{IEEEbiography}
\vspace{11pt}

\begin{IEEEbiography}[{\includegraphics[width=1in,height=1.25in,clip,keepaspectratio]{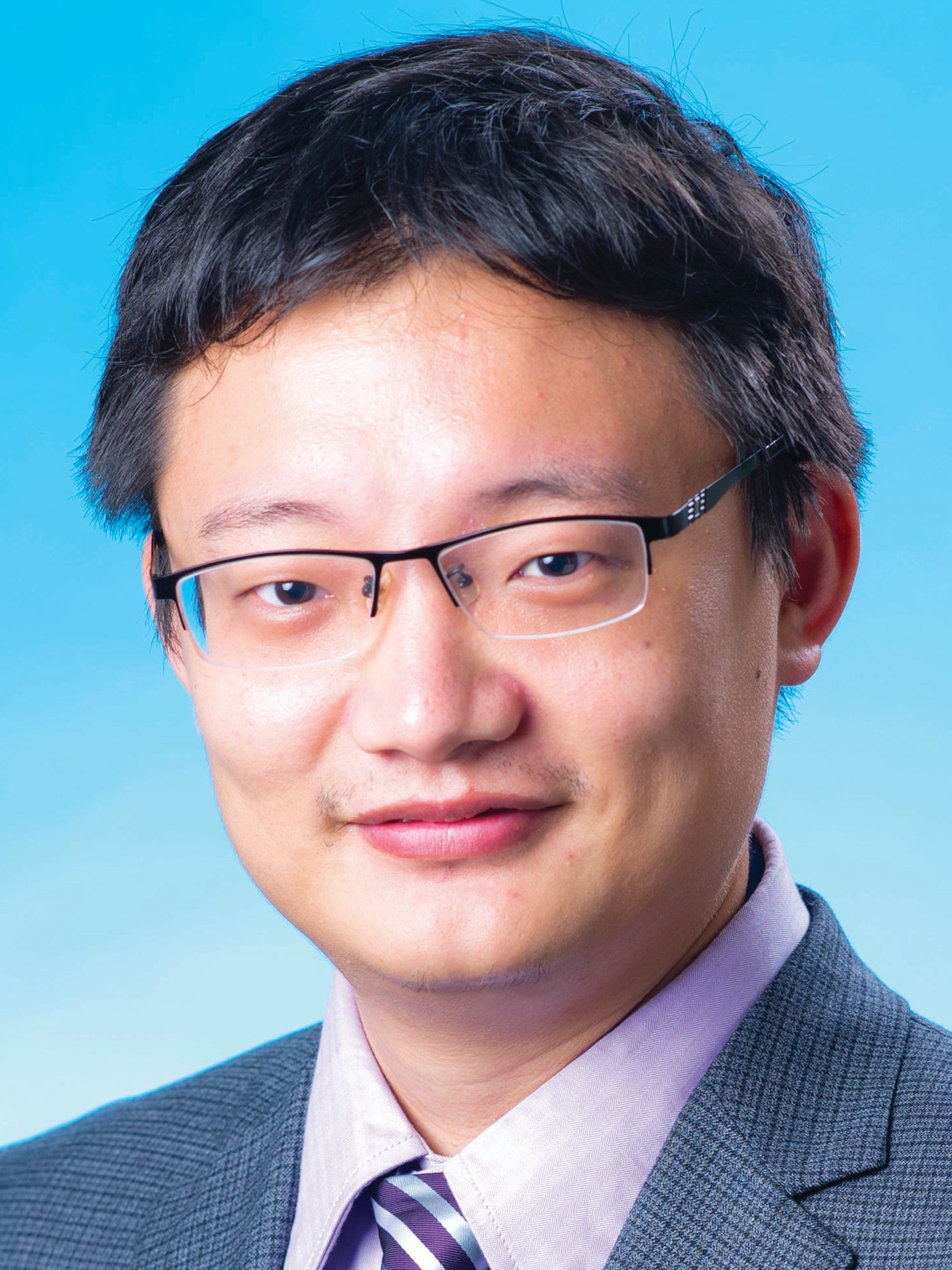}}]{Ming Liu}  (Member 2022)  received the B.A. degree at Tongji University in 2005. He stayed one year in Erlangen-N{\"u}nberg University and Fraunhofer Institute IISB, Germany, as visiting scholar. He graduated as a PhD student from ETH Z{\"u}rich in 2013. He is currently an Assoicate Professor at the Department of Electronic and Computer Engineering, The Hong Kong University of Science and Technology, Hong Kong. He has been involved in several NSF projects, and National 863-Hi-Tech-Plan projects in China. He is PI of 20+ projects including projects funded by RGC, NSFC, ITC, SZSTI, etc. He was the general chair of ICVS 2017, the program chair of IEEE-RCAR 2016, and the program chair of International Robotic Alliance Conference 2017. His current research interests include dynamic environment modeling, 3D mapping, machine learning and visual control, etc.
\end{IEEEbiography}
\vspace{11pt}

\balance
\vfill

\end{document}